\documentclass[11pt,a4paper]{article}
\usepackage[hyperref]{acl2020}
\usepackage{times}
\usepackage{latexsym}

\usepackage{enumitem}
\usepackage{graphicx}
\usepackage{caption}
\usepackage{url}
\usepackage{graphics}
\usepackage{microtype}

\aclfinalcopy % Uncomment this line for the final submission
\title{Prosody Labelled Dataset for Hindi using Semi-Automated Approach}

\author{Esha Banerjee\textsuperscript{1}, Atul Kr. Ojha\textsuperscript{2,3}, Girish Nath Jha\textsuperscript{1}\\
  \textsuperscript{1}Jawaharlal Nehru University, New Delhi,
  \textsuperscript{2}DSI, National University of Ireland Galway, Ireland,\\
  \textsuperscript{3}Panlingua Language Processing LLP, New Delhi\\
   {\tt (esha.jnu,shashwatup9k,girishjjha)@gmail.com}\\}
\date{}

\begin{document}
\maketitle
\begin{abstract}
This study aims to develop a semi-automatically labelled prosody database for Hindi, for enhancing the intonation component in ASR and TTS systems, which is also helpful for building Speech to Speech Machine Translation systems. Although no single standard for prosody labelling exists in Hindi, researchers in the past have employed perceptual and statistical methods in literature to draw inferences about the behaviour of prosody patterns in Hindi. Based on such existing research and largely agreed upon theories of intonation in Hindi, this study attempts to first develop a manually annotated prosodic corpus of Hindi speech data, which is then used for training prediction models for generating automatic prosodic labels. A total of 5,000 sentences (23,500 words) for declarative and interrogative types have been labelled. The accuracy of the trained models for pitch accent, intermediate phrase boundaries and accentual phrase boundaries is 73.40\%, 93.20\%, and 43\% respectively.
\end{abstract}

\section{Introduction}
In order to produce natural sounding speech units, many Automatic Speech Recognition (ASR) and Text-to-Speech (TTS) systems incorporate suprasegmental prosodic features, which generally apply to larger units of representation like phrases or the sentence. Some of the intonational aspects that are covered through prosody include pitch accent, phrasing, duration etc. Spoken in natural rhythm, sentences constitute grammatical breaks and accents which lend specific intonational contours to different sentence types. In general, words may contain lexical stress according to grammatical rules (some words in many languages are not stressed at all), at other times, words may be stressed to convey focus. When strung together, sentences spoken naturally, and impacted by extraneous factors such as speaker motivation, mood, speed etc serve to modify prosodic structure of spoken speech in ways that render it natural sounding to human perception. It is therefore important to be able to input these features, along with the orthography to phonemic conversions, into TTS systems, in order to emulate human-like intelligible voices in building Speech to Speech Machine Translation (SSMT) systems.

Section 2 discusses previous studies in Hindi intonation, with a focused perspective on the development of theories on pitch accent and phrase breaks in Hindi sentences. The theories discussed in this section form the basis for the linguistic analysis and annotation of the declarative and interrogative sentences, discussed in later chapters. Section 3 talks about the speech resource used in the building of this dataset, with a bried overview of the labelling framework discussed in section 4. Sections 5 and 6 discuss the manual and automatic approaches used in the development of this dataset, with the results.

\section{Background}
Hindi belongs to the Indo-European language family and has over 500 million speakers in India. A number of studies exist in literature on Hindi intonation. One of the most pioneer of works was by \cite{moore1965study}, who analyzed Hindi intonation in terms of three different segmental levels in hierarchical relation to each other: foot, measure and sentence. According to his theory, foot consists of one or more syllables in which pitch rises from beginning to end continuously. Measure is the second level of phrasing in which a focused element is separated from the rest of the sentence. Sentence is the topmost level, which encompasses the entire sentence intonation. \cite{harnsberger1994towards} makes an observation along similar lines in which he states that there is a rising pitch contour on content words, in which the low part of the rising contour is a low pitch accent and the high part is either a high trailing tone or boundary tone. The other level of phrasing is the sentence. \cite{FERY2008680} talk about the rising pitch contour on each constituent of the data that they have considered and call it the prosodic phrase and relate it to the syntax of the sentence. \cite{nair2001acoustic}, \cite{dyrud2001hindi} suggest, in their work, that Hindi has lexical stress, such that every word has a particular syllable on which prominence is realized. \cite{sengar2012preliminary}, from their investigative studies, put forward the theory that Hindi is an accentual phrase language and that the Accentual Phrase (AP) was the smallest tonally marked prosodic unit, characterised by a rising contour, the observation being similar to that proposed for Bangladeshi Standard Bengali \cite{khan2008intonational}, a closely related Indo European language. Their research hypothesized that the intonation pattern of Hindi sentences contained a series of APs, characterised by rising contours (which correlate to pitch patterns within the AP) and that the domain of each AP is marked by prosodic boundaries, which may or may not be equal to a single word. The final tone can be overridden by a falling tone in case of declaratives. The entire sentence constituted an IP (intonational phrase) comprised of many ip (intermediate phrase), characterized by silence junctures, and each ip contains one or more APs.

\cite{jyothi2014investigation}, through exploratory investigation using non-expert and expert transcribers, concluded that prosodic phrasing was more consistently agreed on between non-expert transcribers amongst themselves and with the expert transcribers (measured by Cohen’s kappa coefficient). It was also observed that the degree of agreement in prominence (pitch accent) marking was lower, in both cases.

\section{Speech data resource}
The speech corpus obtained and used for this work was developed through the Indian Language Technology Proliferation and Deployment Centre\footnote{\url{http://tdil-dc.in/index.php?option=com_download&task=showresourceDetails&toolid=268&lang=en}} under the Technology Development in Indian Languages (TDIL) program, Ministry of Electronics \& Information Technology (MeitY), MC\&IT, Govt of India. The corpus contains 50 hours of synthetic speech data for both male and female speakers of Hindi. The corpus contains varied sentence kinds (simple, complex) and types (declarative, interrogative, negative, exclamatory etc.) that have been used to choose a varied representation. Sentence units have been selected and extracted from this data for this work.

\section{Labelling framework}
%\begin{enumerate}
%\item
\subsection{Autosegmental-Metrical (AM) model}

The Autosegmental-Metrical framework is a mode of intonational structure that is one of the foremost frameworks used for prosody analysis that was built on the tenets of fundamental work by \cite{pierrehumbert1980phonology} with further refinements by \cite{beckman1986intonational}, \cite{pierrehumbert1988japanese}, \cite{gussenhoven2004phonology} and others. The term ‘autosegmental-metrical’, coined by \cite{ladd2008intonational} was based on the  Autosegmental and Metrical frameworks of phonology, with the autosegmental tier representing intonation structure and metrical tier the phrasing and prominence. Drawn from Autosegmental Phonology, the proposal by \cite{pierrehumbert1980phonology} was that pitch levels are seen as autosegments for intonational analysis while tones are represented by the pitch accent, phrase tone and boundary tone. The tones High (H) and Low (L) were formalized as being associated with stressed syllables as well as prosodic boundaries. The tones associated with stressed syllables were pitch accents and represented with an asterisk (*) while the boundary tones were marked with a percent (\%) sign. In addition, phrasal tones were observed on the intermediate phrase boundaries, which were notated with a hyphen (-). Intermediate phrases were seen to be prosodic units that were larger than the syllable and smaller than the intonational phrase, whose prosodic domain included the whole sentence. Subsequently, this model has been applied to various languages (Japanese \cite{venditti1997japanese}, Korean \cite{jun2000k}, Dutch, German, Italian, French, etc.) with minor modifications. \\

%\item 
\subsection{Tones and Break Indices (ToBI)}

ToBI ( Tones and Break Indices) is a system for transcribing the intonation patterns and other aspects of the prosody of originally, English utterances \cite{beckman1997guidelines}. The labelling scheme consists of:
\begin{itemize}
    \item 6 discrete intonation accents types: H*, !H, L*, L*+H and L+H*.
    \item 2 phrase accent type: H- and L-
    \item 4 boundary tones: L-L\%, L-H\%, H-L\% and H-H\%
    \item 4 break levels: 1, 2, 3, and 4
    \item A HiF0 marker for each intonational phrase
\end{itemize}
An utterance marked using ToBI labeling conventions contains a number of tiers of information: a tone tier, carrying accent information, a break tier for marking prosodic boundaries and a comment tier for miscellaneous information.

ToBI is a standard transcription system for modeling prosodic events of spoken utterances in different languages. It has become a framework to analyze the intonation system and relationship between prosodic and intonation structures of different languages.
%\end{enumerate}

\section{Manual Prosodic Labelling}
500 simple sentences of the types declarative and interrogative were selected and labelled within the frameworks of the intonational framework observed in previously mentioned studies. This annotation follows the proposed framework that the domain of intonation phrase (IP) is the whole utterance, ending with the boundary tone and which may contain one or more intermediate phrases (ip), demarcated by the phrase tone. The smallest prosodic domain is the Accentual Phrase (AP) containing the pitch accent and this may cover one or more words in length. The default pitch accent is observed to be the rising pitch accent (L*Hp) falling on each content word which starts at the left edge of the AP, rising towards the rightmost edge and declines towards the start of the next AP. The only exception is in the final AP, where the boundary tone may override the final AP decline.

Praat\footnote{\url{https://www.fon.hum.uva.nl/praat/}}, a freely available speech analysis software, was used to identify and mark the prosodic boundaries and tones associated with pitch movements. 3 native Hindi speakers with training in phonetics and phonology, transcribed the data.
%\newline
%\begin{enumerate}[label=\Alph*.]
    \subsection{Declarative sentences}
    
\begin{figure}[!ht]
    \centering
    \captionsetup{justification=centering}
    \includegraphics[width=.48\textwidth]{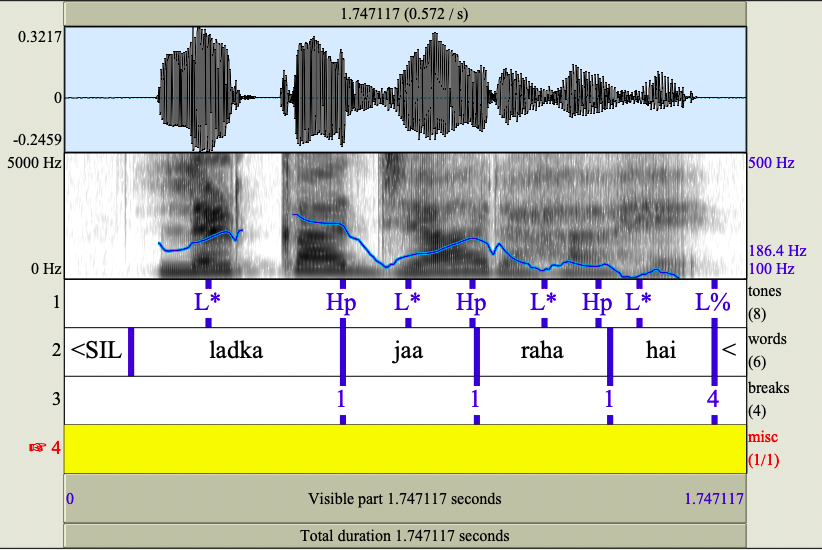}
    \caption{‘ladka ja raha hai’}
    \label{fig:1}
\end{figure}

In Fig~\ref{fig:1}, the declarative sentence is divided into 2 prosodic phrases and shows the pitch pattern L*Hp, overridden by the L\% boundary tone for declaratives.

Figure~\ref{fig:1} example: ladka ja raha hai

\hspace{2.75cm} boy  go is-PROG-MASC

\hspace{2.75cm} The boy is going

This relatively straightforward pattern may be affected by other phenomena that carry information structure, like scrambling and focus. Hindi being a head final, relatively free word order language conveys information by the scrambling of focused constituents to the head of the structure and/or placing a higher pitch accent on the focused element. Focus has also been shown to insert a prosodic break in the post focus word \cite{moore1965study} as well as create a compression in pitch range post focus \cite{harnsberger1996pitch}.
\begin{figure}[ht]
    \centering
    \captionsetup{justification=centering}
    \includegraphics[width=.48\textwidth]{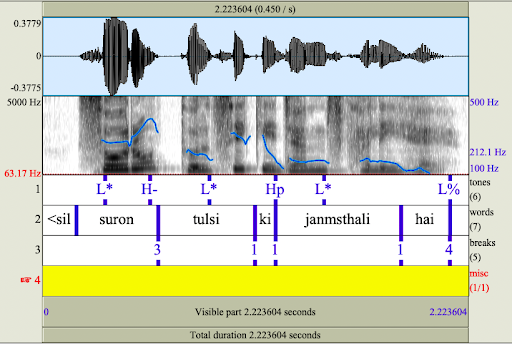}
    \caption{‘Suron Tulsi ki janmasthali hai’}
    \label{fig:2}
\end{figure}

 In Fig~\ref{fig:2} the object ‘suron’ contains the focus and is marked by a relative raised pitch accent compared to the utterance level and the postfocal word is lowered. Figure~\ref{fig:2} example:
 
\hspace{1.75cm}             Suron Tulsi ki janmasthali hai

\hspace{1.75cm}              Suron Tulsi of  birthplace    is

\hspace{1.75cm}               Suron is Tulsi’s birthplace
%\newline
%\item 
%\emph{Interrogative sentences}
\subsection{Interrogative sentences}
\begin{figure}[ht]
    \centering
    \captionsetup{justification=centering}
    \includegraphics[width=.48\textwidth]{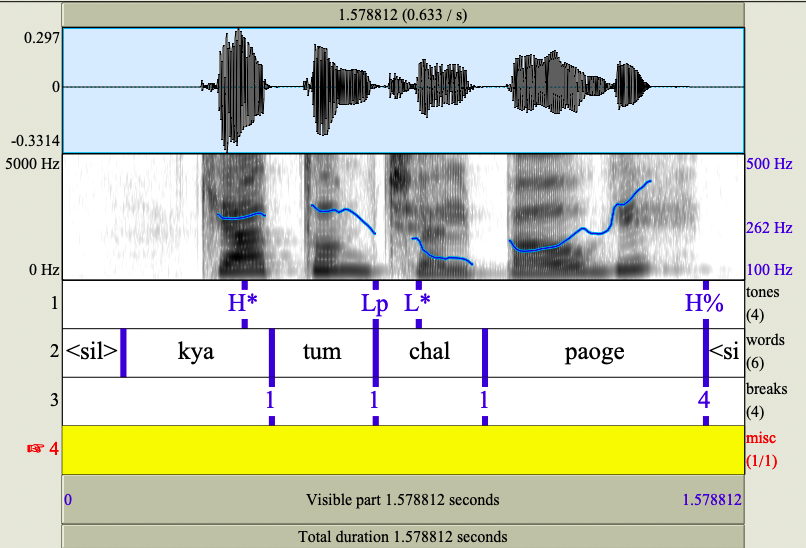}
    \caption{‘kya tum chal paoge’}
    \label{fig:3}
\end{figure}

In the interrogative sentence in Fig~\ref{fig:3}, the intonation pattern follows the L*Hp rising pattern, with a rising H\% boundary tone.

Figure~\ref{fig:3} example: kya tum chal paoge

\hspace{2.75cm}         are you walk able-FUT

\hspace{2.75cm}         will you be able to walk

\begin{figure}[ht]
    \centering
    \captionsetup{justification=centering}
    \includegraphics[width=.48\textwidth]{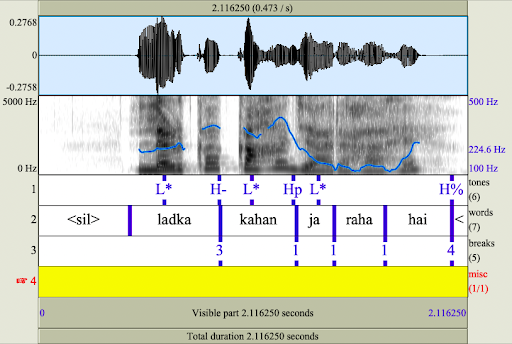}
    \caption{‘ladka kahan ja raha hai’}
    \label{fig:4}
\end{figure}

This was found to be the case in most simple interrogative sentences, except in case of relative higher pitch on seemingly focused elements, as in on the focused ‘kahan’ (where) in Fig~\ref{fig:4}.

Figure~\ref{fig:4} example: ladka kahan ja raha hai

\hspace{2.75cm}         boy where go is-PROG

\hspace{2.75cm}         where is the boy going

The H tone accompanying this rise and fall was observed to have the downtrend component, associated with another closely related Indo-Aryan language, Bengali \cite{jun2014intonational}.

\begin{figure}[ht]
    \centering
    \captionsetup{justification=centering}
    \includegraphics[width=.47\textwidth]{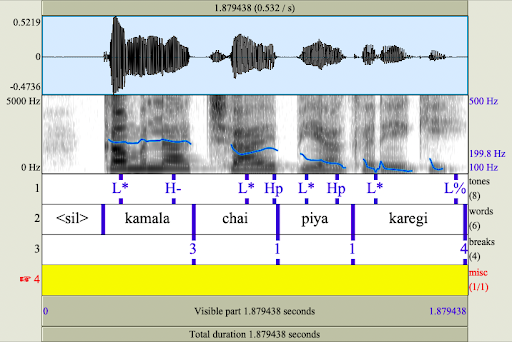}
    \caption{‘Kamala chai piya karegi’}
    \label{fig:5}
\end{figure}

Figure~\ref{fig:5} example: Kamala chai piya karegi

\hspace{2.75cm} Kamala tea drink do-HABI

\hspace{2.75cm} Kamala will drink tea

The downtrend observed in the consecutive H tones in Fig~\ref{fig:5} are consistent with the observation that H tones in successive APs are of a lower pitch than the preceding. 
Apart from minor effects of microprosody, there were not many deviations observed in this intonation pattern

%\end{enumerate}

\subsection{Inter-Annotator Agreement}
Three linguists (native speakers of Standard Hindi) belonging to Delhi, with familiarity in ToBI annotation conventions, were asked to label the dataset. Some initial training was provided for the analysis as well as the annotation labels presented to them for this research. Initial training consisted of individual instructions as well as calculating inter annotator consistency, and this was carried out iteratively to achieve the desired accuracy. Overall transcriber agreement (calculated using Cohen's kappa) for prosodic breaks was 0.87, and for pitch accents was 0.69.

\section{Automatic Prosodic Labelling}
The manually labelled sentences developed in the previous section has been used as training data to fine-tune Au-ToBI, an existing automatic prosody labelling toolkit widely available \cite{rosenberg2010autobi}, by building newly trained models within their standard specifications. The study is a comparative analysis on the performance of accuracy between pre-existing and newly trained Au-ToBI models for this research. Au-ToBI was particularly selected for its adaptability to ToBI, which had been used as the labelling conventions for the manually annotated data as well.

\subsection{Automatic ToBI}
Au-ToBI~\citep{rosenberg2010autobi} is a publicly available tool that runs on Java, which contains models trained on English sentences to automatically detect and extract prosodic breaks and pitch accents from spoken utterances . Based on pre-trained models of English, initial detection of pitch accents and phrase boundaries is carried out, based on cues like pitch excursions and silence duration. This is followed by the classification of the phrase boundary tones and type of pitch accent prediction. The classification of prosodic breaks and pitch accents is done as per the ToBI annotation conventions.

\subsection{Experiments in Automatic Labelling}
This experiment was conducted in two parts. The Hindi manual prosodic dataset developed in section 5 was divided into training and test data in 90:10 ratio. The first experiment on the English model was evaluated with the test data, while the second experiment was conducted with the training and test sets. The experiments are divided into two steps. First, use pre-trained English model detection and classification algorithms in Au-ToBI to generate automatic labels for Hindi utterances and measure accuracy, and second, use manual annotated data to build Hindi prosody models for Au-ToBI.

The pre-processing of this dataset consisted of manually segmentation of sentences into words in the TextGrid files. The transcription was carried out in Devanagari. Since the Hp boundary tone for AP was a distinct feature from the standard ToBI guidelines, the tones in the training sentences were mapped to their corresponding ToBI labels. The “breaks” section in the uploaded files was converted to Au-ToBI format, under the alignment process. This included conversion of “number” to “time” etc. TextGrid and WAV files were named similarly and located in the same folder for use in the training.

Parameters and values for all three tiers “words”, “breaks” and “tones” were implemented, along with the Hindi model classifiers and detectors. Multiple command lines were provided for training pitch accent, intonational phrase boundary, intermediate phrase boundary, phrase accent and boundary tone detection and classification models. The default features were selected for the building of these models,  using feature extractor and feature classifier. Since the test files used for prediction of Hindi labels came from one speaker, normalization parameters were not used in this set up. The built Hindi Au-Tobi models were evaluated on the test data. The 50 hours TDIL speech corpus was used to extract a further 4,500 declarative and interrogative sentences, split 50:50 for declarative and interrogative sentences.

\subsection{Results}
Results are output as TextGrid files and in addition to the “words” tier that was present, contains two additional generated tiers named “tones” (for generated pitch accents) and “breaks” (for generated prosodic breaks). The accuracy of the models are demonstrated in the below figure.
\begin{figure}[ht]
    \centering
    \captionsetup{justification=centering}
    \includegraphics[width=.51\textwidth]{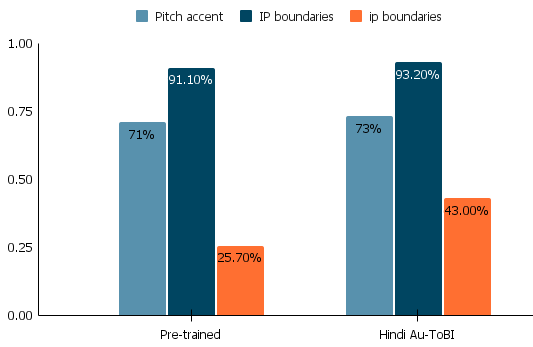}
    \caption{‘Results of pre-trained and newly trained Au-ToBI models’}
    \label{results}
\end{figure}

\section{Conclusion and Future Work}
The sentences labelled with the prosodic labels are a valuable source of training data for ASR and TTS systems to introduce the naturalness component that is often derived from prosodic elements. This study aims to employ the studies on intonational behavior of simple declarative and interrogative sentences in Hindi done in recent years and develop a semi-automatically annotated labelled dataset that can be used to enhance the prosodic output in SSMT systems for a natural sounding voice. The approach is modeled on the principles of Tones and Break Indices (ToBI) annotation guidelines, and recent research on prosodic boundaries and prominence marking in Hindi and related languages. 2,550 words (500 sentences) are manually annotated and these sentences are used to extend the corpus size up to 5,000 sentences, using Automatic ToBI \cite{rosenberg2010autobi}, \cite{jyothi2014investigation}. The research aims to develop a prosody labelled database for Hindi for training speech models for natural sounding voices. The prosodic labelled dataset and developed Hindi-AuToBi model will be available on GitHub at \url{https://github.com/esha-banerjee/Hindi_Au-ToBI}.

\section*{Acknowledgments}
Atul Kr. Ojha would like to acknowledge the EU’s Horizon 2020 Research and Innovation programme through the ELEXIS project under grant agreement No. 731015.
\bibliography{anthology,acl2020}
\bibliographystyle{acl_natbib}

\end{document}